\pdfoutput=1
\documentclass[11pt]{article}

\usepackage[preprint]{acl}

\usepackage{times}
\usepackage{latexsym}

\usepackage[T1]{fontenc}
\usepackage[utf8]{inputenc}

\usepackage{microtype}

\usepackage{inconsolata}

\usepackage{graphicx}
\usepackage{enumitem}

\usepackage{hyperref}
\usepackage{url}

\usepackage{ebproof}
\usepackage{caption}
\usepackage{subcaption}
\usepackage{booktabs}
\usepackage{wrapfig}
\usepackage{algpseudocode}
\usepackage{algorithm}

\usepackage{amsmath}
\usepackage{amssymb}
\usepackage{mathtools}
\usepackage{amsthm}
\usepackage{float}
\usepackage{multirow}
\usepackage{subcaption}
\usepackage{listings}

\newcommand{\trans}[0]{\intercal}

\title{The Rotary Position Embedding May Cause Dimension Inefficiency \\in Attention Heads for Long-Distance Retrieval}

\author{Ting-Rui Chiang \\
  University of Southern California \\
  \texttt{tingruic@usc.edu} \\\And
  Dani Yogatama \\
  University of Southern California \\
  \texttt{yogatama@usc.edu} \\}

\begin{document}
\maketitle
\begin{abstract}
The Rotary Position Embedding (RoPE) is widely used in the attention heads of many large language models (LLM).
It rotates dimensions in the query and the key vectors by different angles according to their positions in the input sequence.
For long context modeling, the range of positions may vary a lot, and thus RoPE rotates some dimensions by a great range of angles.
We hypothesize that the wide range of rotation angles may prevent LLMs from utilizing those dimensions.
To validate this hypothesis, we present a controlled experiment showing that applying RoPE causes low utility of certain dimensions.
Our analyses on three LLMs also indicate that these dimensions do not help LLMs do long-context question answering.
% This finding suggests directions for improving LLMs.
\end{abstract}

\section{Introduction}

\citet{roformer} proposed the Rotary Position Embedding (RoPE) for Transformer models~\citep{transformers}.
Because it is parameter-free and computationally efficient, it has been widely adopted in many large language models (LLMs), such as PaLM~\citep{chowdhery2022palm}, Gemma~\citep{team2024gemma,team2024gemma2}, LLaMA~\citep{touvron2023llama,touvron2023llama2,dubey2024llama}, OLMo~\citep{groeneveld-etal-2024-olmo,olmo20242olmo2furious}, Mistral~\citep{jiang2023mistral}, Falcon~\citep{almazrouei2023falcon}, and Qwen~\citep{bai2023qwen,qwen2.5}. 
Despite the success of these LLMs on several downstream tasks, LLMs have also been found to be less effective when handling longer context~\citep{liu-etal-2024-lost,nih,an-etal-2024-l,bai-etal-2024-longbench,zhang-etal-2024-bench,li2024long}.

Most recent work has focused on understanding and mitigating LLM failure to generalize to long context.
For example, \citet{kazemnejad2023the} inspected different positional encoding methods.
\citet{han-etal-2024-lm} and \citet{xiao2024efficient} explained the failure with the distribution shifts of LLMs' internal representation.
\citet{an2025why} attributed the failure to the skewed length distribution in the training data.
\citet{peng2024yarn}, \citet{ntk}, and \citet{ntk2} addressed the failure by studying ways to interpolate RoPE.

Orthogonal to existing studies, our work analyzes the impact of RoPE on models' utilization of dimensions in attention heads.
We hypothesize that, for long distance attention, the way that RoPE rotates the query and the key vectors may prevent the model from utilizing the dimensions that it rotates significantly. 
Our results of a controlled experiment and analyses of three real-world large language models support our hypothesis.

As RoPE has been widely used in many LLMs, our findings have great implications.
Not utilizing certain dimensions means that the computational cost for those dimensions may not be necessary.
LLMs may be made more computationally efficient by pruning these dimensions.
It also implies that LLMs may achieve better performance on long-context tasks with the same number of parameters if they utilize more dimensions.
Addressing this issue is thus paramount for LLM developers.

\section{Background: Rotary Position Embeddings (RoPE)}

\citet{roformer} proposed the Rotary Position Embedding (RoPE), which can be applied to the key and query vectors for attention operations.
It encodes relative position by rotating the intermediate representations according to their positions in the input sequence.
Specifically, RoPE rotates a vector in $\mathbb{R}^{2D}$ at position $m$ with a block-diagonal matrix
\begin{equation}
\begin{split}
&M^{(D)}_m =
\begin{bmatrix} 
M_{m\theta_1} & & & & \\
%& M_{m\theta_2} & & & \\
& & \ddots & & \\
% & & & M_{m\theta_{D - 1}} & \\
& & & & M_{m\theta_{D}} \\
\end{bmatrix},\\
&\text{with } M_{m\theta_i} = \begin{bmatrix}
    \cos{m\theta_i} & -\sin{m\theta_i} \\
    \sin{m\theta_i} & \cos{m\theta_i}
\end{bmatrix},
\end{split}
\label{eq:rope-matrix}
\end{equation}
for some scalars $\theta_1 > \theta_2 > \cdots \theta_D$ that decide the frequency of the rotations.

Let a query vector and a key vector at position $m, n$ be $q_m, k_n \in \mathbb{R}^{2D}$.
RoPE rotates them with RoPE matrices $M_m$ and $M_n$ respectively, so the dot product for computing the attention weight is
\begin{equation}
    \begin{split}
        &\mathrm{RoPE}(q_m) \cdot \mathrm{RoPE}(k_n) \\ 
        &= (M_m q_m)^\trans (M_n k_n) = q_m^\trans (M_{n - m}) k_n.
    \end{split}
    \label{eq:rope-dot}
\end{equation}
\iffalse
\citet{roformer} suggested that this design causes the inner product to decay as the relative position increases.

\paragraph{Two Observations} 
Based on this descriptions, we can have two apparent observations:
\begin{enumerate}
    \item From Eq.~\ref{eq:rope-dot}, we can see that RoPE rotates query and key vectors with angles computed based on their relative position.
    \item From Eq.~\ref{eq:rope-matrix}, we can see that RoPE rotates the first few dimensions with higher frequencies $\theta_1, \theta_2 \cdots$ and the last few dimensions slowly with lower frequencies $\theta_D, \theta_{D-1}, \cdots$.
\end{enumerate}
\fi

\section{Dimension Inefficiency}
\label{sec:hypothesis}

We hypothesize that RoPE may cause dimension inefficiency in attention heads for long-dependency modeling.
Specifically, when a task requires an attention head to attend to a distant position, RoPE may prevent the attention from utilizing the first few dimensions in its attention heads.
This is because RoPE rotates those dimensions with greater rates ($\theta$'s in Eq.~\ref{eq:rope-matrix})\footnote{In practice, the dimensions are reordered for computational efficiency. Here we assume that the order of the dimensions is by the magnitude of $\theta$, from greater to smaller.}.
For long-context tasks, such as long-context question answering, the possible relative positions $m - n$ between a key vector $k_n$ for the target information and a query vector $q_m$ can vary greatly, so the rotation applied on these dimensions can be \textit{any} angles.
Therefore, the model cannot produce query and key vectors such that their first few dimensions can consistently contribute a positive value to the inner product in Eq.~\ref{eq:rope-dot}.
We hypothesize that the first few dimensions are thus useless for long-distance attention.

\section{Controlled Experiment}
\label{sec:toy}

\begin{figure}
    \centering
    \begin{subfigure}[t]{0.23\textwidth}
         \includegraphics[width=\textwidth]{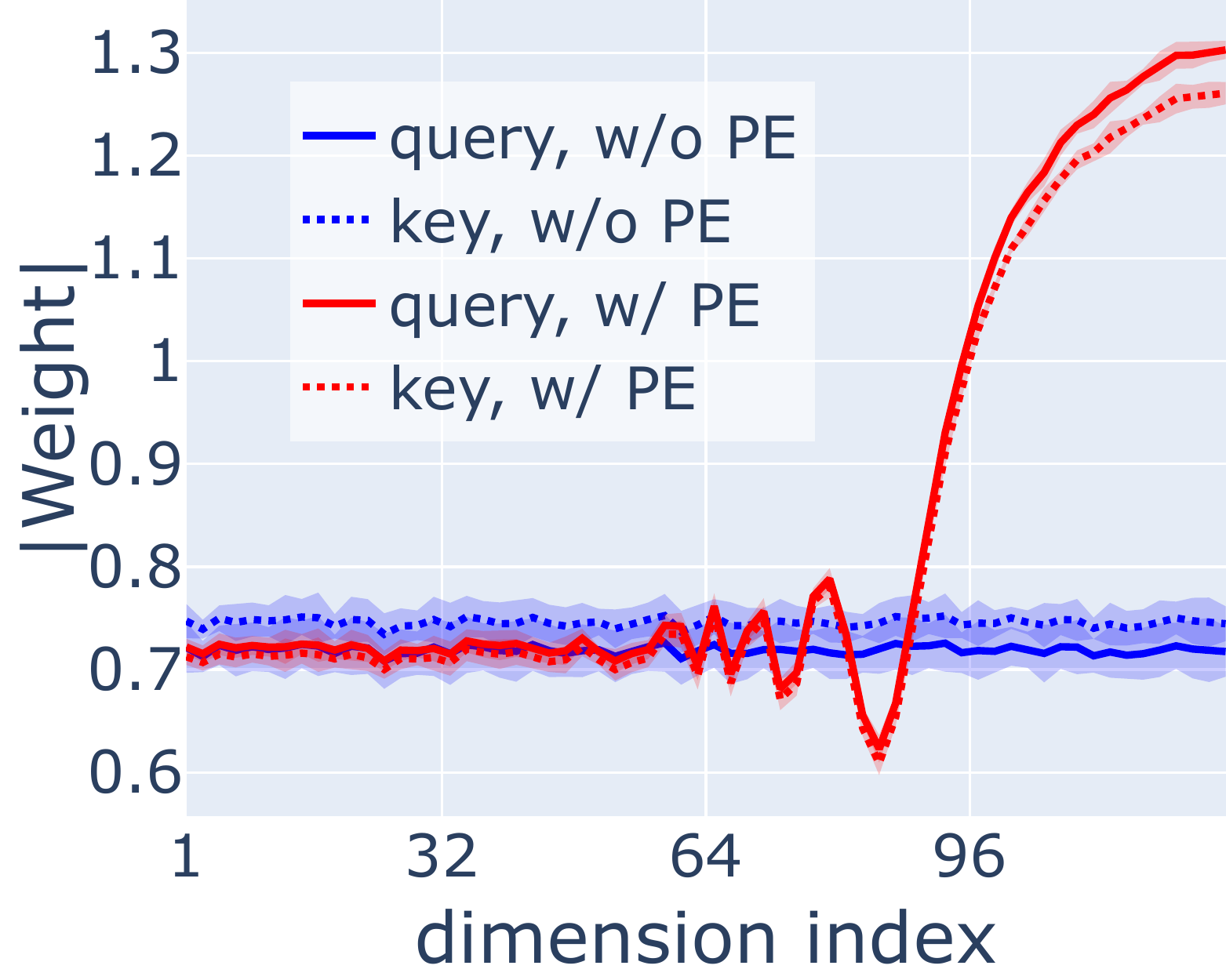}
         \caption{The average magnitude of the key and the query vectors for each dimension.}
         \label{fig:toy-exp-weight}
    \end{subfigure}  
    \hfill
    \begin{subfigure}[t]{0.23\textwidth}
         \includegraphics[width=\textwidth]{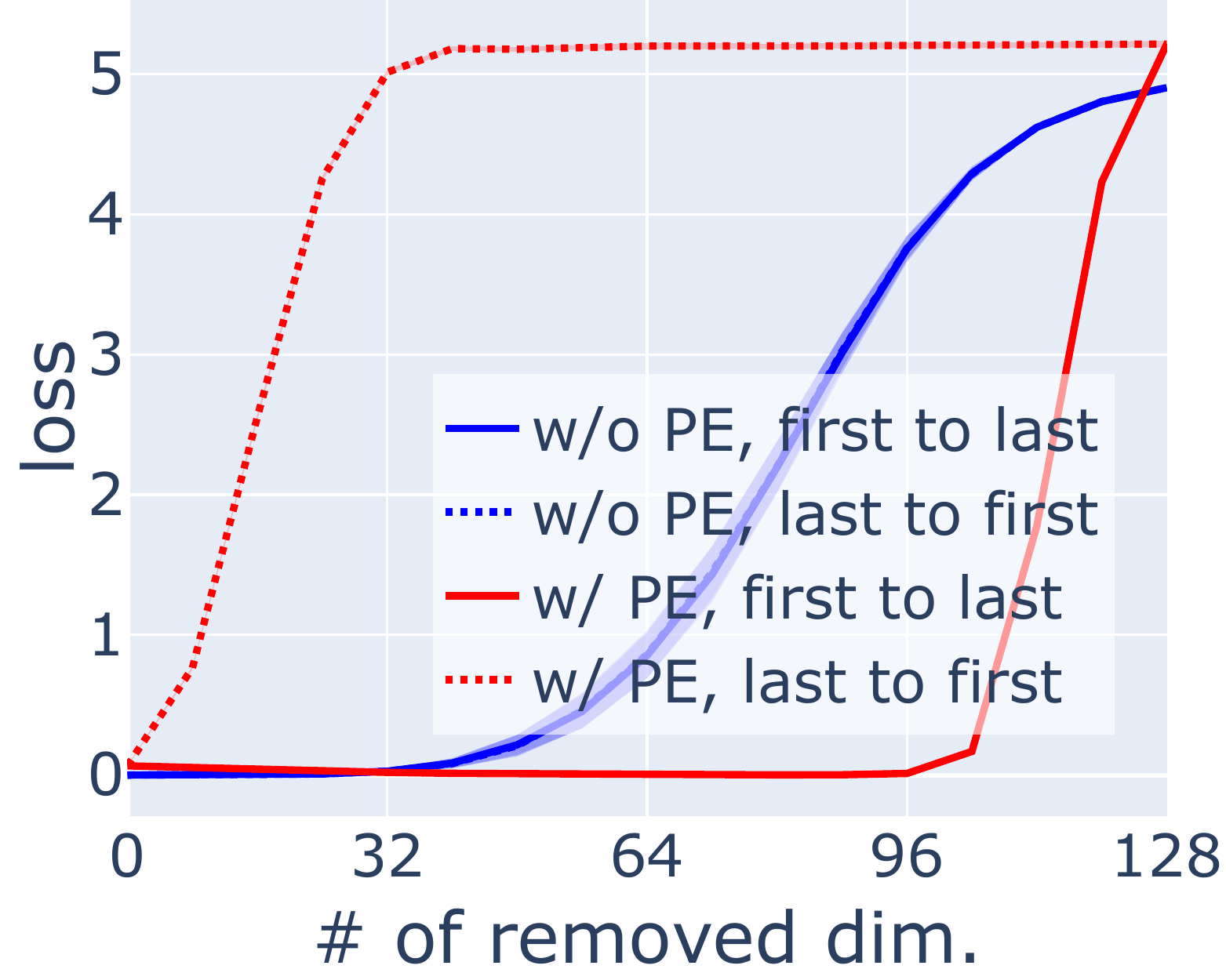}
         \caption{\# of removed dimensions v.s. loss in Eq.~\ref{eq:toy-loss} ($-\mathbb{E} \log P(v_i|q_i, K, V)$).}
         \label{fig:toy-exp-loss}
    \end{subfigure} 
    \caption{Analysis of the dimensions in the attention head of the models (w/ and w/o applying RoPE) in \S\ref{sec:toy}.}
    \label{fig:toy-exp}
\end{figure}

We present a controlled experiment to demonstrate how RoPE can cause dimension inefficiency.
We design a simple experiment where the model needs to learn $n$ vector tuples $\{ (q_i, k_i ,v_i) \}^n_{i=1}$ such that the attention head can retrieve $v_i$ with $q_i$ from any randomly sampled subset of key-value pairs $\{ (k, v) | k \in K, v \in V \} \subset \{ (k_i ,v_i) \}^n_{i=1}$.
Specifically, we optimize the following objective function:
\begin{equation}
\begin{split}
    &\min_{\{q_i, k_i v_i\}_{i=1}^n} - \frac{1}{n} \sum_{i=1}^n \mathbb{E}_{K, V} \log P( v_i | q_i, K , V) \\
    &\text{where } P(v_i | q_i) = \frac{\exp{(a^\trans v_i)}}{\sum_{j}^n \exp(a^\trans v_j)}, \\
    &a = \mathrm{Attention}(q_i, K, V).
\end{split}
\label{eq:toy-loss}
\end{equation}
We train models in two setups, one with RoPE applied on $K$ and the other without (details in \S\ref{sec:toy-details}).

\paragraph{Results and Discussion} 
Our experimental results indicate that RoPE causes dimension inefficiency.
Firstly, we plot the average weight of $\{q\}_{i=1}^n$ and $\{k\}_{i=1}^n$ for each dimension in Figure~\ref{fig:toy-exp-weight}.
It shows that the model trained with RoPE applied learns to assign lower weights to the first few dimensions of $\{q_i\}_{i=1}^n$ and $\{k_i\}_{i=1}^n$.
Secondly, Figure~\ref{fig:toy-exp-loss} shows that, when RoPE is applied, removing the first few dimensions does not affect the loss significantly, while removing the last few dimensions greatly increases the loss.
This indicates that the model relies mainly on the last few dimensions and does not utilize the first few dimensions.
In contrast, the models without RoPE do not exhibit these phenomena.
This is in line with our hypothesis in \S\ref{sec:hypothesis}.

\section{Inspecting Real-world Models}
\label{sec:llm}

\begin{figure}
    \centering
    \begin{subfigure}[t]{0.23\textwidth}
         \includegraphics[width=\textwidth]{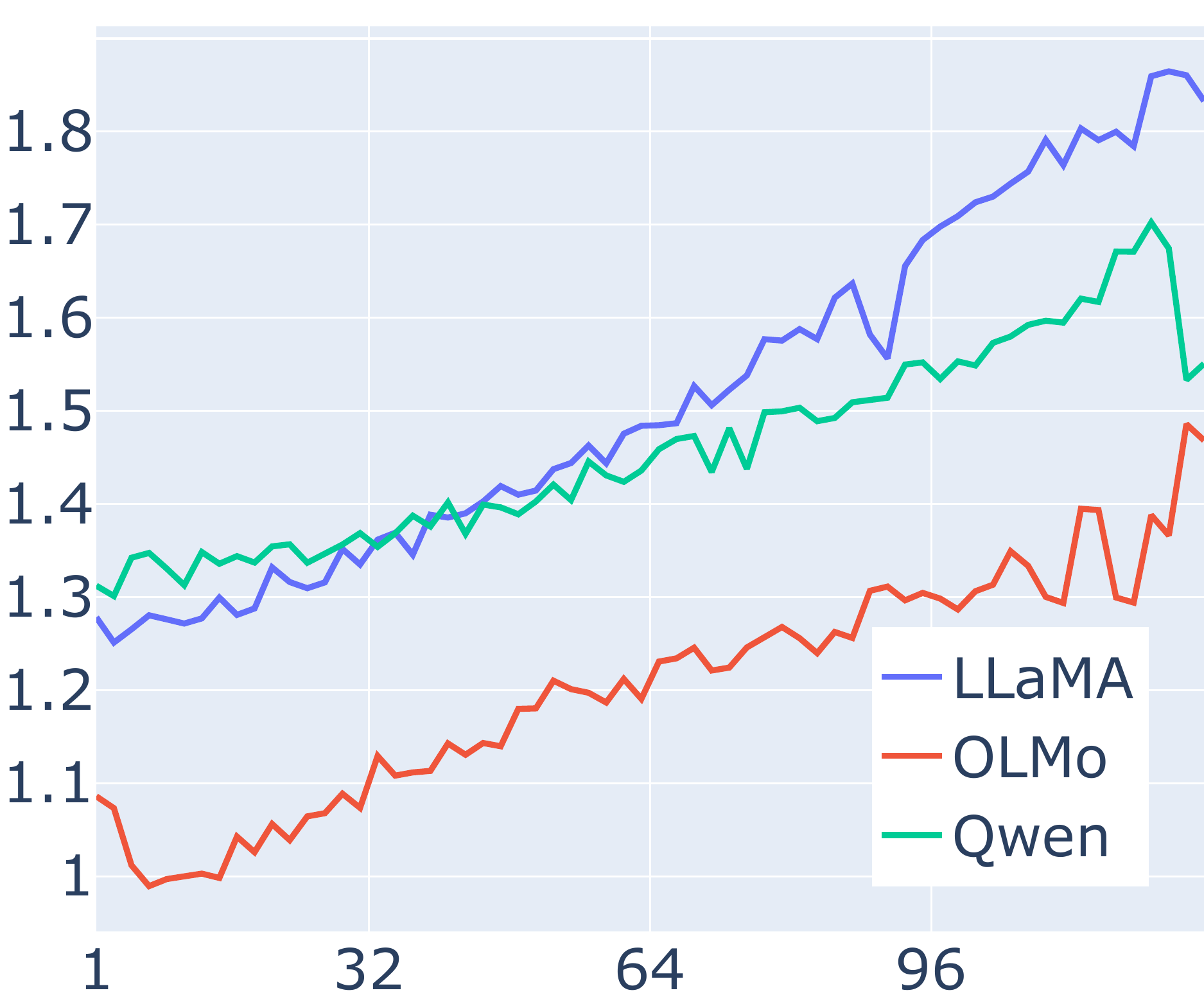}
         \caption{Average L1 norm}
         \label{fig:llm-attn-weight}
    \end{subfigure}
    \hfill
    \begin{subfigure}[t]{0.23\textwidth}
         \includegraphics[width=\textwidth]{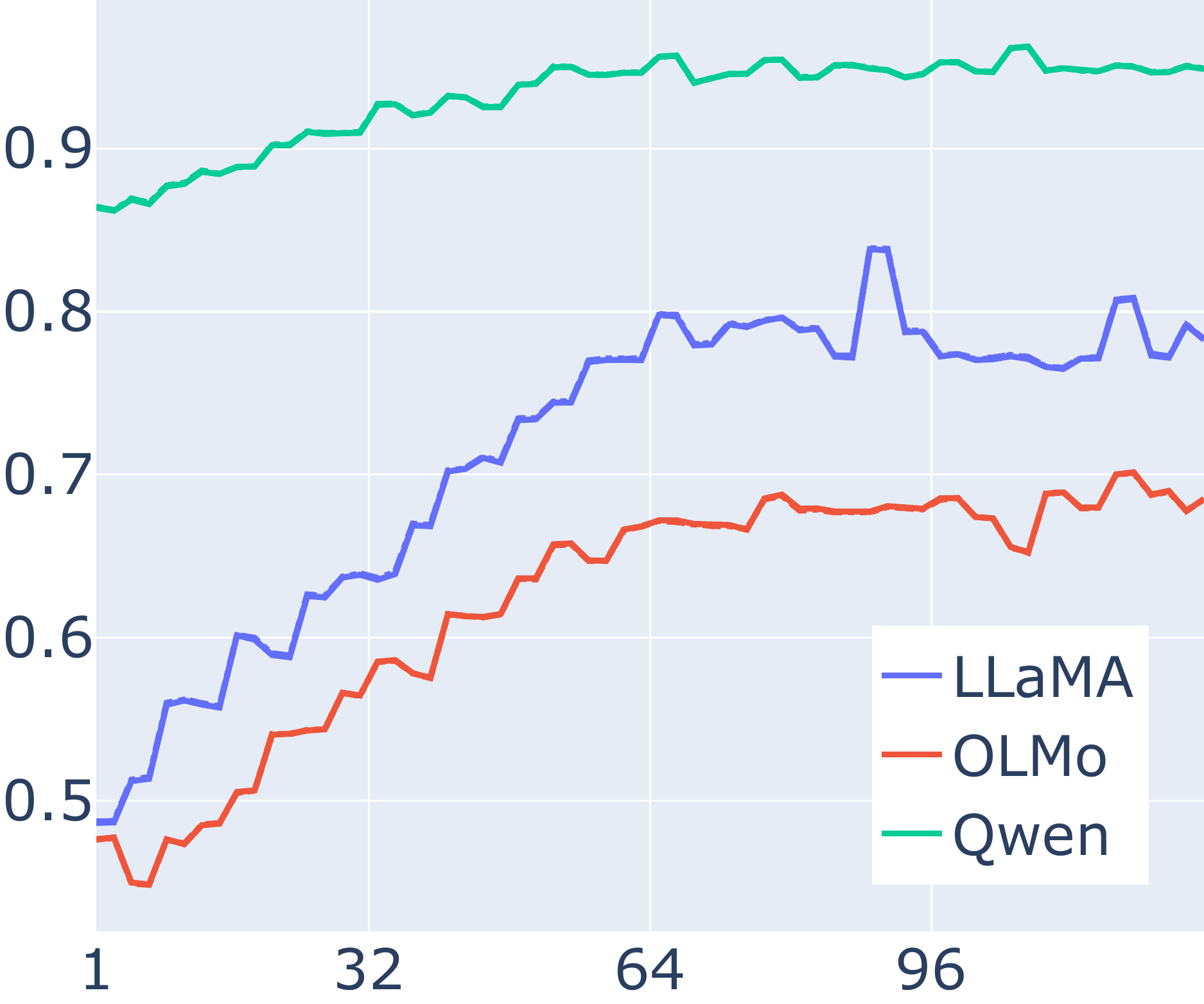}
         \caption{Average utility score.}
         \label{fig:llm-attn-utility}
    \end{subfigure}
    \caption{The average importance of each dimensions in the query vectors of the attention heads, measured by the L1 norm of rows in the query weight matrices (left) and by utility score in \S\ref{sec:utilization} (right).
    We visualize all heads in Figure~\ref{fig:all-weight} and Figure~\ref{fig:all-mask}.
    }
    \label{fig:llm-attn-weights}
\end{figure}

We then inspect three 7B/8B large language models (LLM), Llama-3.1-8B-Instruct~\citep{dubey2024llama}, QWen-2.5-7B-Instruct~\citep{qwen2.5}, and OLMo-2-7B-Instruct~\citep{olmo20242olmo2furious}.
These models have 128 dimensions in their attention heads.
For quick inspection, we first plot the L1 norm of the rows in the query projection matrices in all the attention heads in Figure~\ref{fig:llm-attn-weights}.
It shows increasing trends for all three models, as we have found in the toy experiment (Figure~\ref{fig:toy-exp-weight}). 
As the L1 norm of the rows may not directly reflect the importance of the dimensions in the query vectors, we utilize a long-context task for further investigations.

\subsection{Experimental Setup}

As we hypothesize that the dimension inefficiency only occurs for attention heads that model long dependency, we choose a task that involves long dependence modeling, the long-context question-answering task.
We follow the setup of~\citet{liu-etal-2024-lost}, where we provide the model with 20 documents for each question, among which only one contains the answer.
Following \citet{liu-etal-2024-lost}, we measure the accuracy for scenarios where the answer is in the 1st, 10th, and 20th document.

\begin{table}[]
\centering
\begin{tabular}{lccccc}
\toprule
      &     LLaMA & OLMo & Qwen \\
\midrule
Original  & 54.09 & 56.66 & 58.72 \\
Masked    & 54.15 & 57.55 & 57.36 \\
\bottomrule
\end{tabular}
\caption{The performance of LLMs before and after masking dimensions with low utility. We average the accuracy of setups where the answer is in the 1st, 10th, 20th document (full results in Table~\ref{table:mask-full}).}
\label{table:utility}
\end{table}

\subsection{Utilization of Dimensions}
\label{sec:utilization}

\paragraph{Identifying dimension utilization} 
To identify the dimensions in the query vectors that are not crucial to attention, we train a sparse mask that masks out as many dimensions as possible while preserving the attention head's output.
Specifically, for each attention head of $2D$ dimensions at layer $\ell$ with index $i$, we find a masking vector $u_{\ell,i} \in [0, 1]^{2D}$ over the query vector that minimizes
\begin{align}
    &\| \mathrm{Attn}_{\ell,i}(q, K, V) - \mathrm{Attn}_{\ell,i}(q \odot u_{\ell,i}, K, V) \|^2_2 \nonumber \\ 
    &+ \alpha \| u_{\ell,i} \|_1,
    \label{eq:utilization}
\end{align}
where we set the hyper-parameter $\alpha = \frac{1}{2D}$.
We then treat the value of $u$ in each dimension as the utilization score of that dimension.

\paragraph{Experiment}
We first prompt the LLM to answer the questions in the dataset.
Then we feed in the LLM the concatenation of the instruction, the documents, the question, and LLMs' generation, optimizing Eq.~\ref{eq:utilization}.
% he result will reflect the utilization of the dimensions in the generation process.

\paragraph{Sanity Check}
To check whether the dimensions with low utility scores ($u$ in Eq.~\ref{eq:utilization}) are indeed not crucial for the LLM, we conduct a sanity check.
For each head at layer $\ell$ with index $i$, based on $u_{\ell,i}$, we mask dimensions whose utility scores is less than 0.5 and prompt the model to answer the same questions again.
We then measure the performance of the masked model.
Table~\ref{table:utility} shows that masking these dimensions does not greatly decrease performance, suggesting that dimensions with low scores are not crucial to performance.

\paragraph{Observations and Discussion}
We also plot the average utility score of each dimension ($u_{\ell, i}$ averaged over $\ell$'s and $i$'s).
Figure~\ref{fig:llm-attn-utility} shows increasing trends for the three models, indicating a lower utility of the first few dimensions.
This is also in line with our hypothesis.

\begin{figure}
    \centering
     \begin{subfigure}[b]{0.23\textwidth}
         \centering
         \includegraphics[width=\textwidth]{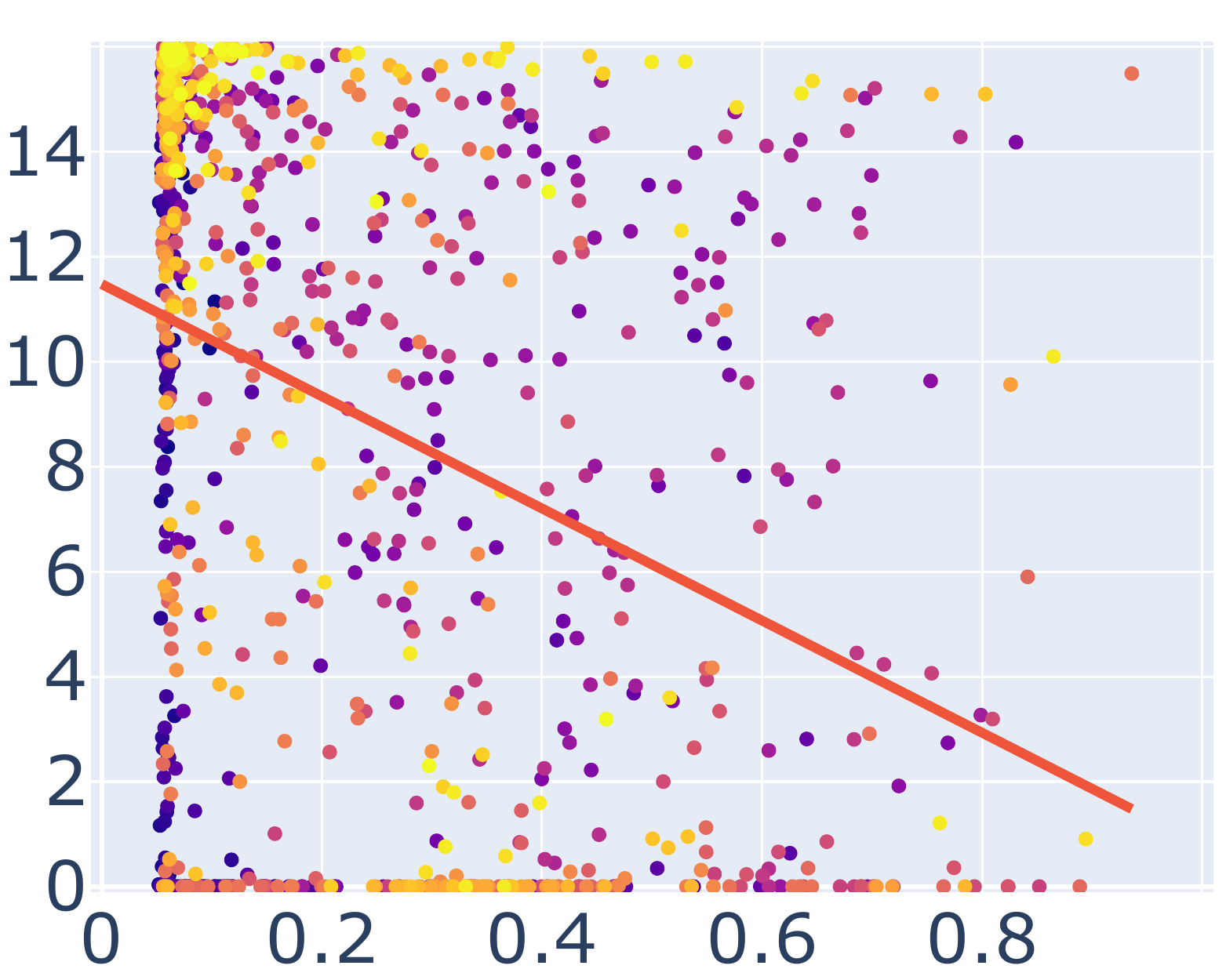}
         \caption{LLaMA first, $\rho=-0.34$}
         \label{fig:y equals x}
     \end{subfigure}    
     \begin{subfigure}[b]{0.23\textwidth}
         \centering
         \includegraphics[width=\textwidth]{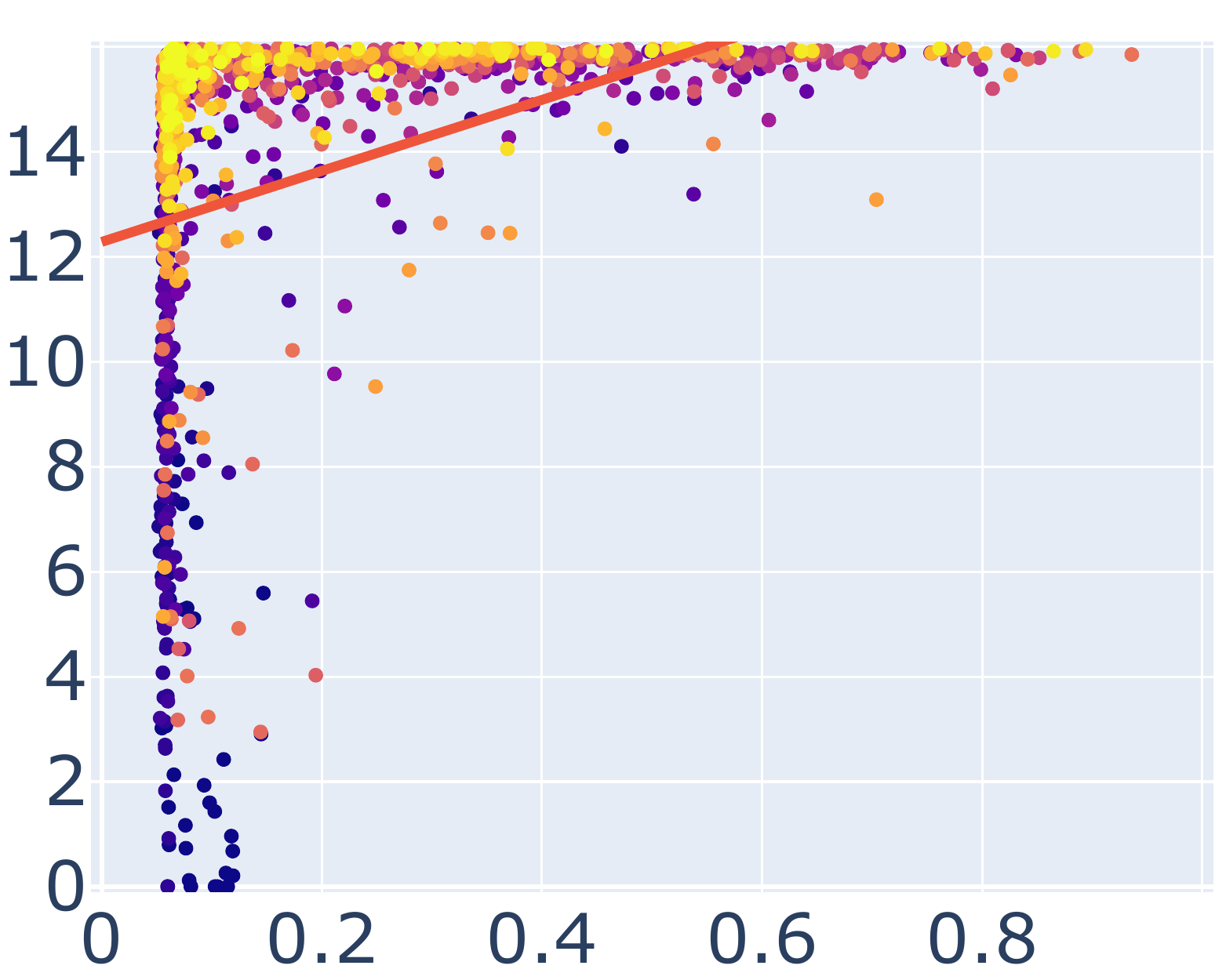}
         \caption{LLaMA last, $\rho=0.38$}
         \label{fig:y equals x}
     \end{subfigure}    
     \begin{subfigure}[b]{0.23\textwidth}
         \centering
         \includegraphics[width=\textwidth]{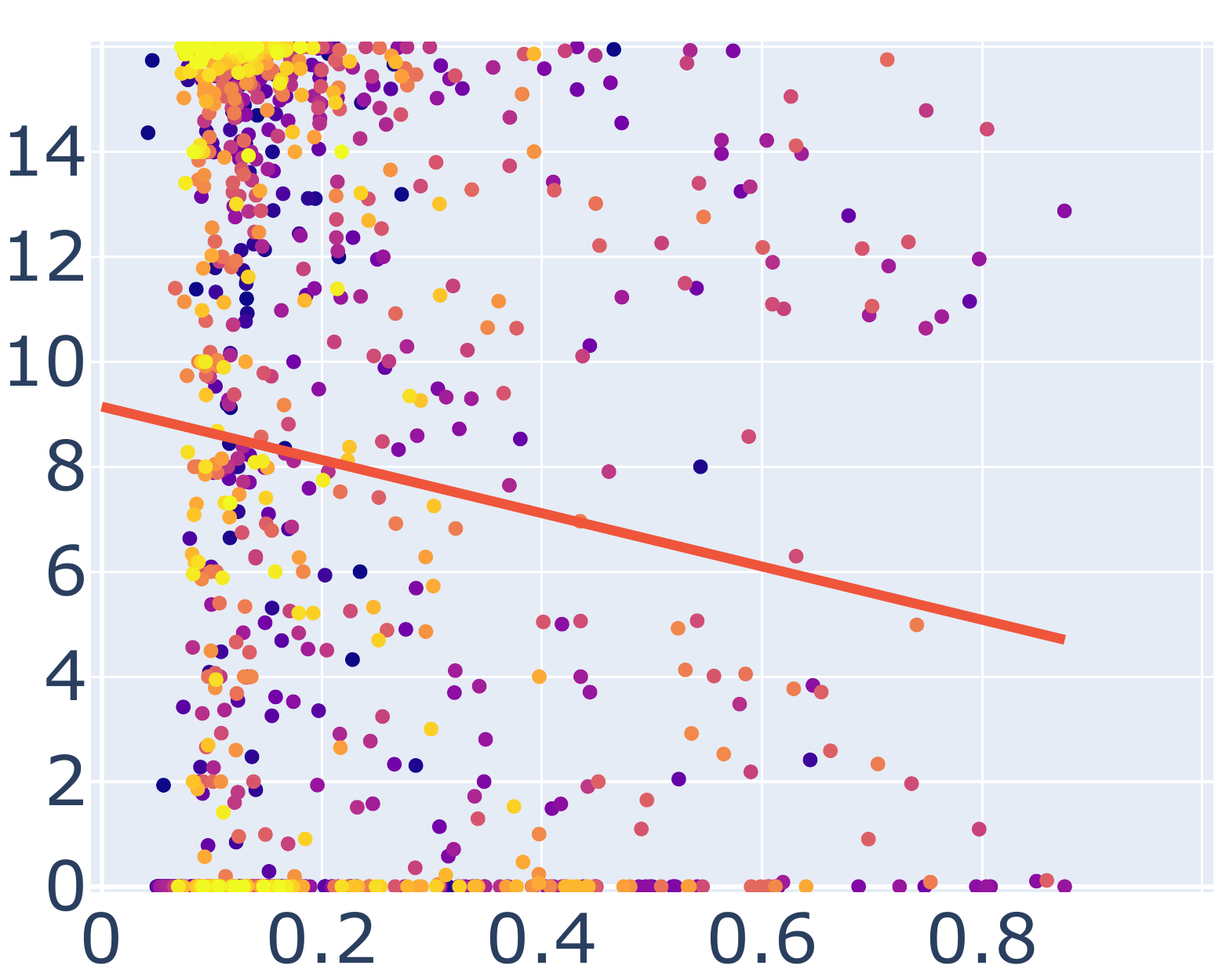}
         \caption{OLMo first, $\rho=-0.12$}
         \label{fig:y equals x}
     \end{subfigure}    
     \begin{subfigure}[b]{0.23\textwidth}
         \centering
         \includegraphics[width=\textwidth]{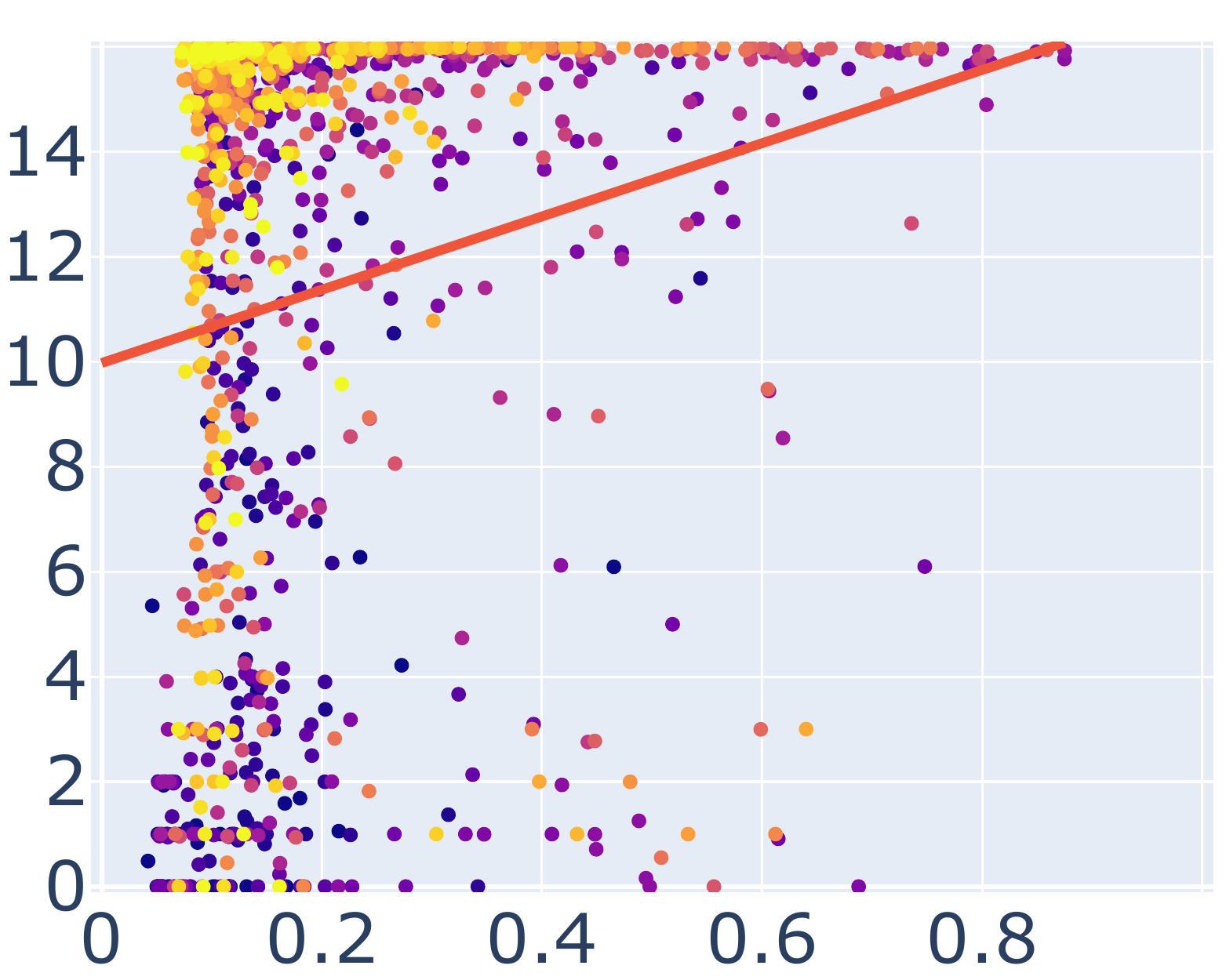}
         \caption{OLMo last, $\rho=0.20$}
         \label{fig:y equals x}
     \end{subfigure}    
     \begin{subfigure}[b]{0.23\textwidth}
         \centering
         \includegraphics[width=\textwidth]{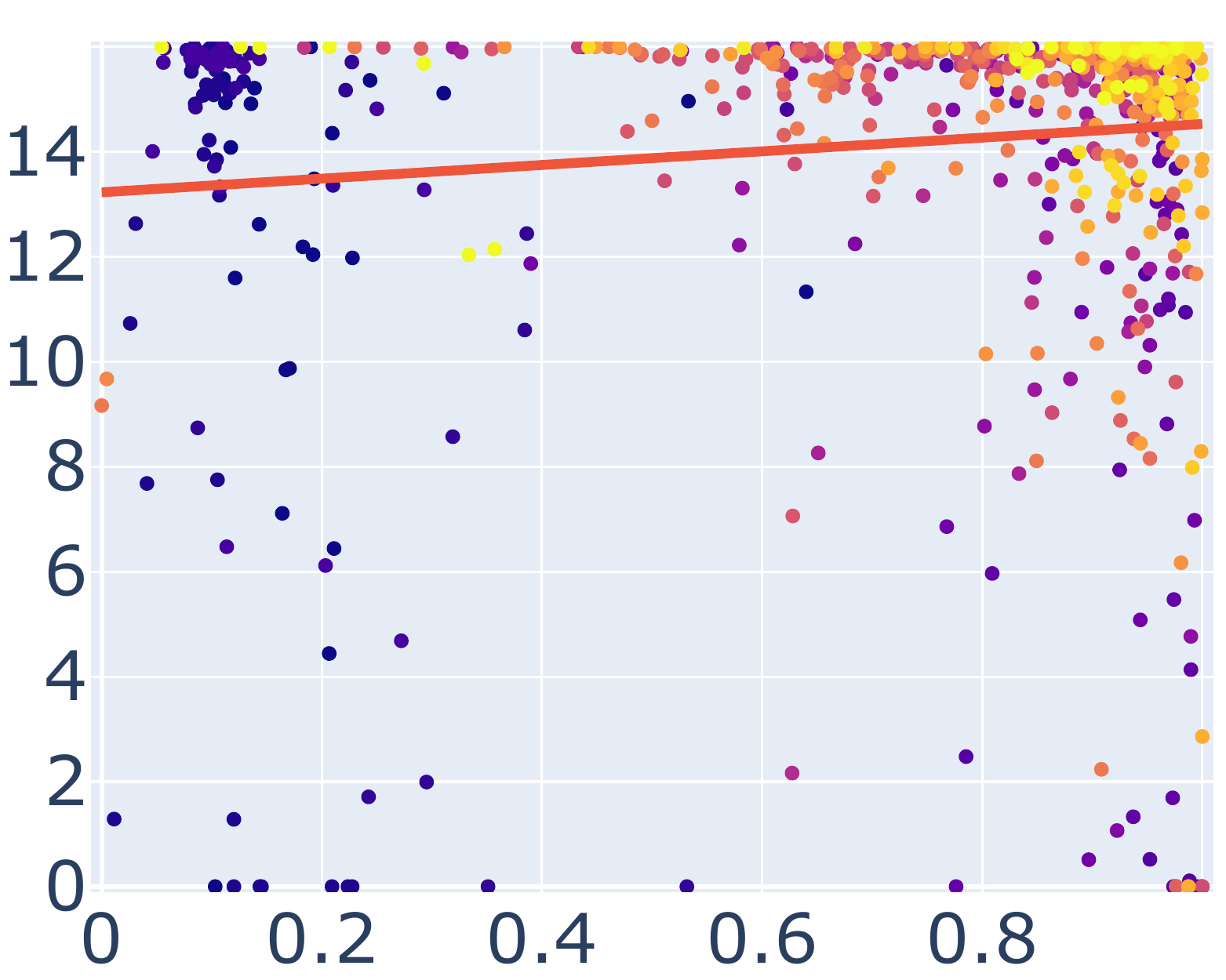}
         \caption{Qwen first, $\rho=0.1$}
         \label{fig:y equals x}
     \end{subfigure}    
     \begin{subfigure}[b]{0.23\textwidth}
         \centering
         \includegraphics[width=\textwidth]{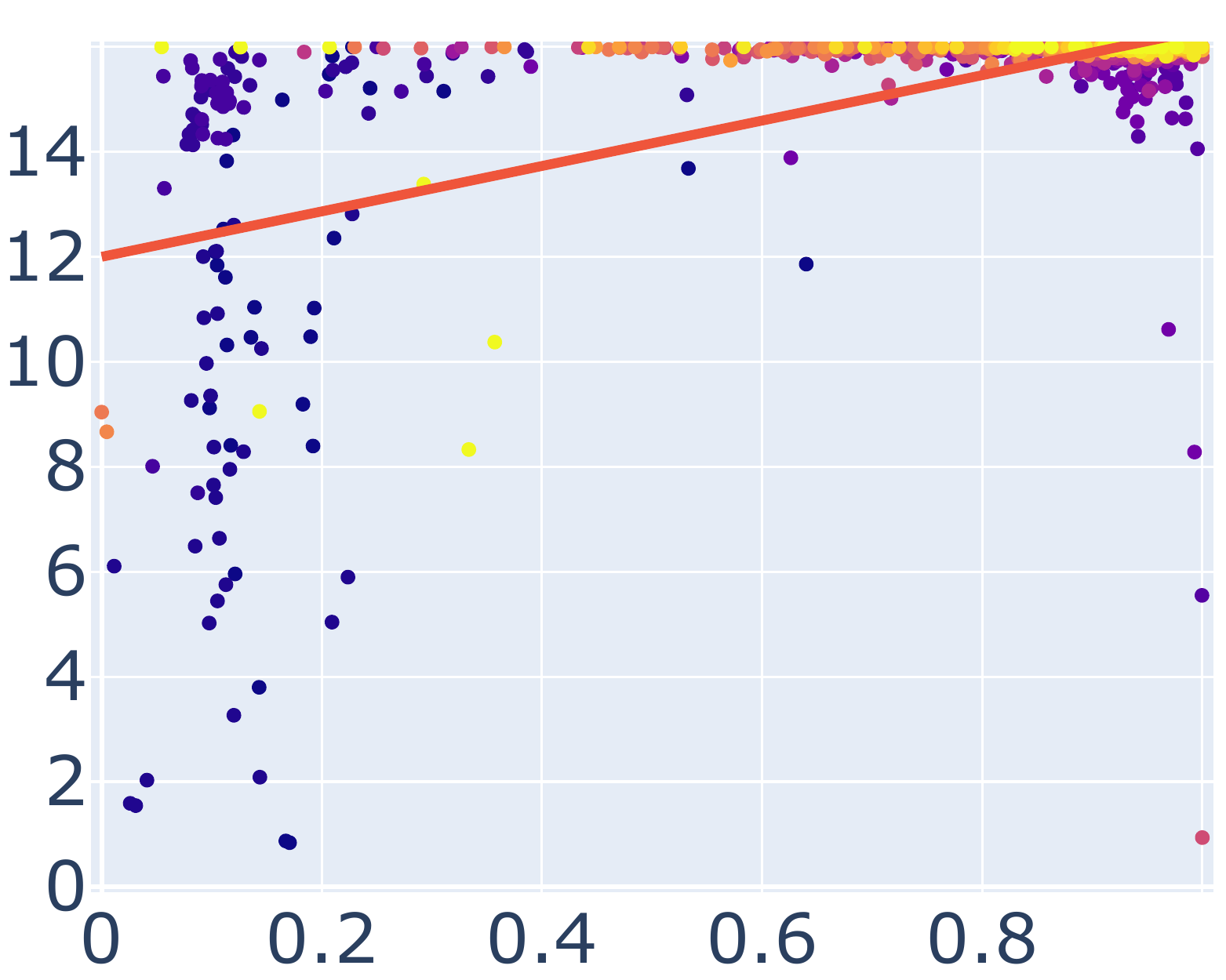}
         \caption{Qwen last, $\rho=0.54$.}
         \label{fig:y equals x}
     \end{subfigure}        
    \caption{The relationship between the retrieval-head indicator score (x-axis) and utility score of the first 16 or last dimensions (y-axis). Each dot represents an attention head. The lighter dot color represents the deeper layers. The red line represents the linear regressor.}
    \label{fig:attn-vs-utility}
\end{figure}

\subsection{Retrieval Heads vs Dimension Inefficiency}
\label{ref:retrieval-head}

We then inspect whether heads for long-distance attention rely less on the first few dimensions.
According to \citet{xiao2024duoattention} and \citet{wu2025retrieval}, LLMs tend to have a small subset of attention heads, called \textit{retrieval heads}, that are responsible for retrieving information from long context, while the majority of the heads are \textit{streaming heads}, dedicated to modeling local context. 
In this section, we examine the dimension inefficiency of these retrieval heads.

\paragraph{Identifying Retrieval Heads}
We use the statistics of LLMs' attention scores over the context to identify retrieval heads.
Specifically, for each head, we measure the sum of the attention weight between the context part (instruction and documents).
\footnote{We ignore the attention weight on the begin-of-string token because \citet{xiao2024efficient} suggest that models tend to use it as an \textit{sinkhole}.} 
and the question-output part and use the sum as a retrieval-head indication score.
Compared with the approach by~\citet{xiao2024duoattention}, our method does not require gradient computation and thus is faster.
% Additionally, our approach is more fine-grained, as we operate at the head-level, while \citet{xiao2024duoattention}

\begin{table}[]
\centering
\begin{tabular}{lc|cccc}
\toprule
                       &           & 1st     & 10th    & 20th   & Avg \\
                       \midrule
\multirow{5}{*}{Llama} & $\phi$ & 60{\small.49} & 53{\small.18} & 48{\small.59} & 54{\small.09} \\
                       & [:16]   & 60{\small.79} & 53{\small.75} & 50{\small.32} & 54{\small.95} \\
                       & [:32]  & 59{\small.51} & 52{\small.77} & 47{\small.88} & 53{\small.39} \\
                       & [-16:]  & 13{\small.82} & 17{\small.29} & 51{\small.98} & 27{\small.70} \\
                       & [-32:]  & 4{\small.07} & 5{\small.27} & 36{\small.20} & 15{\small.18} \\ \hline
\multirow{5}{*}{OLMo}  & $\phi$ &  59{\small.32} & 53{\small.45} & 57{\small.21} & 56{\small.66} \\
                       & [:16]   & 60{\small.19} & 52{\small.92} & 57{\small.48} & 56{\small.86} \\
                       & [:32]  & 60{\small.83} & 52{\small.84} & 57{\small.02} & 56{\small.90} \\
                       & [-16:]  & 42{\small.21} & 40{\small.11} & 56{\small.12} & 46{\small.15} \\
                       & [-32:]  & 29{\small.87} & 33{\small.37} & 46{\small.48} & 36{\small.57} \\ \hline
\multirow{5}{*}{Qwen}  & $\phi$ &  60{\small.41} & 57{\small.14} & 58{\small.61} & 58{\small.72} \\ 
                       & [:16]   & 63{\small.81} & 58{\small.98} & 61{\small.13} & 61{\small.31} \\
                       & [:32]  & 60{\small.72} & 51{\small.22} & 57{\small.66} & 56{\small.53} \\
                       & [-16:]  & 19{\small.62} & 20{\small.60} & 19{\small.51} & 19{\small.91} \\
                       & [-32:]  & 0{\small.30} & 0{\small.45} & 0{\small.98} & 0{\small.58} \\    
                       \bottomrule
\end{tabular}
\caption{The performance of LLMs when the first or last $n$ (denoted as [:$n$] or [$-n$:]) out of 128 dimensions in the retrieval-heads are masked. $\phi$ means no dimensions are masked.
The columns are for the setup where the answer is in the 1st, 10th, 20th document.
}
\label{tab:mask}

\end{table}

\paragraph{Observation and Discussion}
We plot the relationship between the retrieval head indication scores and the utility score of the first or last 16 dimensions in Figure~\ref{fig:attn-vs-utility}.
There are positive (Pearson) correlations between the utility of the last few dimensions and the retrieval head indicator scores. 
The last few dimensions in the retrieval heads also generally have higher utility scores.
For LLaMA and OLMo, we also see that the first few dimensions in the retrieval heads tend to have lower utility scores, while Qwen is an exception, which may be due to a caveat of the utility score.
We discuss more in the next section.

\subsection{Causal Intervention on Retrieval Heads' Dimensions}

Although the experiments in \S\ref{sec:utilization} provide us with a macro view over all the attention heads, the measurement of dimension utility in \S\ref{sec:utilization} has caveats.
The utility scores only indicates which dimensions affect the intermediate representations more, but do not distinguish what causes the LLM to generate the correct answer. 
Dimensions with high utility score may be even harmful for the LLM's performance.
% In our toy experiment, we observe that the loss drop slightly
% The LLM may be sensitive to these dimensions because it has not learned to ignore them.
A more direct way to inspect the effect of some dimensions would be masking those dimensions and evaluating the performance.

\paragraph{Experimental Setup}
We inspect whether the first few dimensions are, as suggested by our hypothesis, not helpful for the model to generate correct answers.
To do so, we inspect the effect of masking dimensions in the attention heads whose retrieval indication scores are greater than 0.5.
We measure the performance when the first 16, 32 or the last 16, 32 dimensions out of 128 dimensions are masked.

\paragraph{Results and Discussion}
Table~\ref{tab:mask} shows the effect of masking dimensions in the attention heads.
It shows that masking the first 16 dimensions slightly increases the average performance.
Masking the first 32 dimensions decreases the average accuracy by less than 2.2\%.
These results indicate that the first few dimensions, as suggested by our hypothesis, do not help the model produce the correct answer.
Masking the last 32 dimensions, in contrast, is detrimental to the LLMs' performance.
Masking the last 16 dimensions also hurts the performance, but it hurts less when the correct answer is in the last (20th) documents, i.e., the document closest to the question.
It suggests the last few dimensions are crucial for long-distance attention, but are less crucial when the distance is shorter.

\section{Conclusion}

In this work, we hypothesize that the Rotary Position Embedding (RoPE) may prevent LMs from utilizing all the dimension for long-context modeling.
We also provide supporting evidence, including a toy experiment \S\ref{sec:toy}, and a deep inspection of three LLMs \S\ref{sec:llm}.
Based on our finding, we suggest that future LLM creators consider alternatives of RoPE, or at least not use RoPE for all attention heads.

\section{Limitations}

One limitation of our work is that, due to limited computational resources, we experiment with only three 7B/8B LLMs. However, given the consistent results across these models, we believe our findings generalize to other LLMs using RoPE.
Additionally, while \citet{liu-etal-2024-lost} also evaluate LLMs using a key-value retrieval task, we focus on long-context question answering, which we consider a more realistic setting.
Finally, our primary goal is to raise awareness of RoPE’s potential issues and encourage further research. 
We do not explore how our findings could improve LLMs, such as enhancing computational efficiency.
We also leave the mitigation of the dimensional deficiency for future work, as it may require significant computational resource for additional fine-tuning.

\section*{Acknowledgements}

We appreciate Joshua Robinson, Xinyan Velocity Yu, Ollie Liu for providing valuable comments on this paper. 

\bibliography{custom}

\appendix

\section{Details of the Controlled Experiment in \S\ref{sec:toy}}
\label{sec:toy-details}

We train attention models with 128 hidden dimensions.
We sample 128 out of 1000 key-value pairs for the $K, V$ in Eq.~\ref{eq:toy-loss}. 
We use a learning rate of 1e-3, a batch size of 64, a maximum position of 2048, 10000 samples per epoch and train the model for 100 epochs. 
The implementation and configuration of RoPE are the same as the one for LLaMA.

% \section{Details of Retrieval Head Identification}
% When computing the retrieval indicator scores, we treat the instruction, documents as the context, the question and the LLM's generation as the query.
% For each attention head, we 

\section{Prompts Templates}

We prompt LLaMA with the following template

\begin{lstlisting}[breaklines]
<|start_header_id|>system<|end_header_id|>

Write a high-quality answer for the given question using only the provided search results (some of which might be irrelevant).
<|eot_id|><|start_header_id|>user<|end_header_id|>

Document [1](Title: {title}) {content}
Document [2](Title: {title}) {content}
Document [3](Title: {title}) {content}
...

Question: {question}
<|eot_id|><|start_header_id|>assistant<|end_header_id|>
\end{lstlisting}

We prompt OLMo with the following template. 

\begin{lstlisting}[breaklines]
<|endoftext|><|user|>

Write a high-quality answer for the given question using only the provided search results (some of which might be irrelevant).

Document [1](Title: {title}) {content}
Document [2](Title: {title}) {content}
Document [3](Title: {title}) {content}
...

Question: {question}
<|assistant|>
\end{lstlisting}

We prompt Qwen with the following template. 

\begin{lstlisting}[breaklines=true]
<|im_start|>system
Write a high-quality answer for the given question using only the provided search results (some of which might be irrelevant).<|im_end|>
<|im_start|>user

Document [1](Title: {title}) {content}
Document [2](Title: {title}) {content}
Document [3](Title: {title}) {content}
...

Question: {question}
<|im_end|>
<|im_start|>assistant
\end{lstlisting}

\section{Dataset}

We use the processed dataset from \citet{liu-etal-2024-lost}.
They released it under the MIT license.
It is derived from NaturalQuestions-Open~\citep{kwiatkowski-etal-2019-natural,lee-etal-2019-latent}.
It can be downloaded at \url{https://github.com/nelson-liu/lost-in-the-middle/tree/main/qa_data/20_total_documents}.
The language is English.
There are 2655 examples in the test set.

\section{Computational Resource}
We conduct each experiment with one NVIDIA RTX A6000 GPU.
Generating answers for one setup takes about 3-5 hours.
Collecting attention statistics and computing the utility score takes about 45 minutes per setup, respectively.

\section{Package Version}

We use the following Python packages:
\begin{itemize}
    \item torch: 2.5.1
    \item transformers: 4.48.2
    \item numpy: 2.0.2
\end{itemize}

\begin{table}[]
\begin{tabular}{lccccc}
\toprule
      &          & 1st & 10th & 20th & Avg. \\
\midrule
\multirow{ 2}{*}{Llama} & original & 60{\small.49} & 53{\small.18} & 48{\small.59} & 54{\small.09} \\
                        & masked   & 58{\small.98} & 52{\small.66} & 50{\small.81} & 54{\small.15} \\ \hline
\multirow{ 2}{*}{OLMo}  & original  & 59{\small.32} & 53{\small.45} & 57{\small.21} & 56{\small.66} \\
                        & masked   & 59{\small.51} & 53{\small.52} & 59{\small.62} & 57{\small.55} \\ \hline
\multirow{ 2}{*}{Qwen}  & original  & 60{\small.41} & 57{\small.14} & 58{\small.61} & 58{\small.72} \\
                        & masked   & 60{\small.49} & 55{\small.25} & 56{\small.35} & 57{\small.36} \\
\bottomrule
\end{tabular}
\caption{The detailed performance of LLMs before and after masking dimensions with low utility when the answer is in the 1st, 10th, 20th document.}
\label{table:mask-full}
\end{table}

\begin{figure}
    \centering
     \begin{subfigure}[b]{0.23\textwidth}
         \centering
         \includegraphics[width=\textwidth]{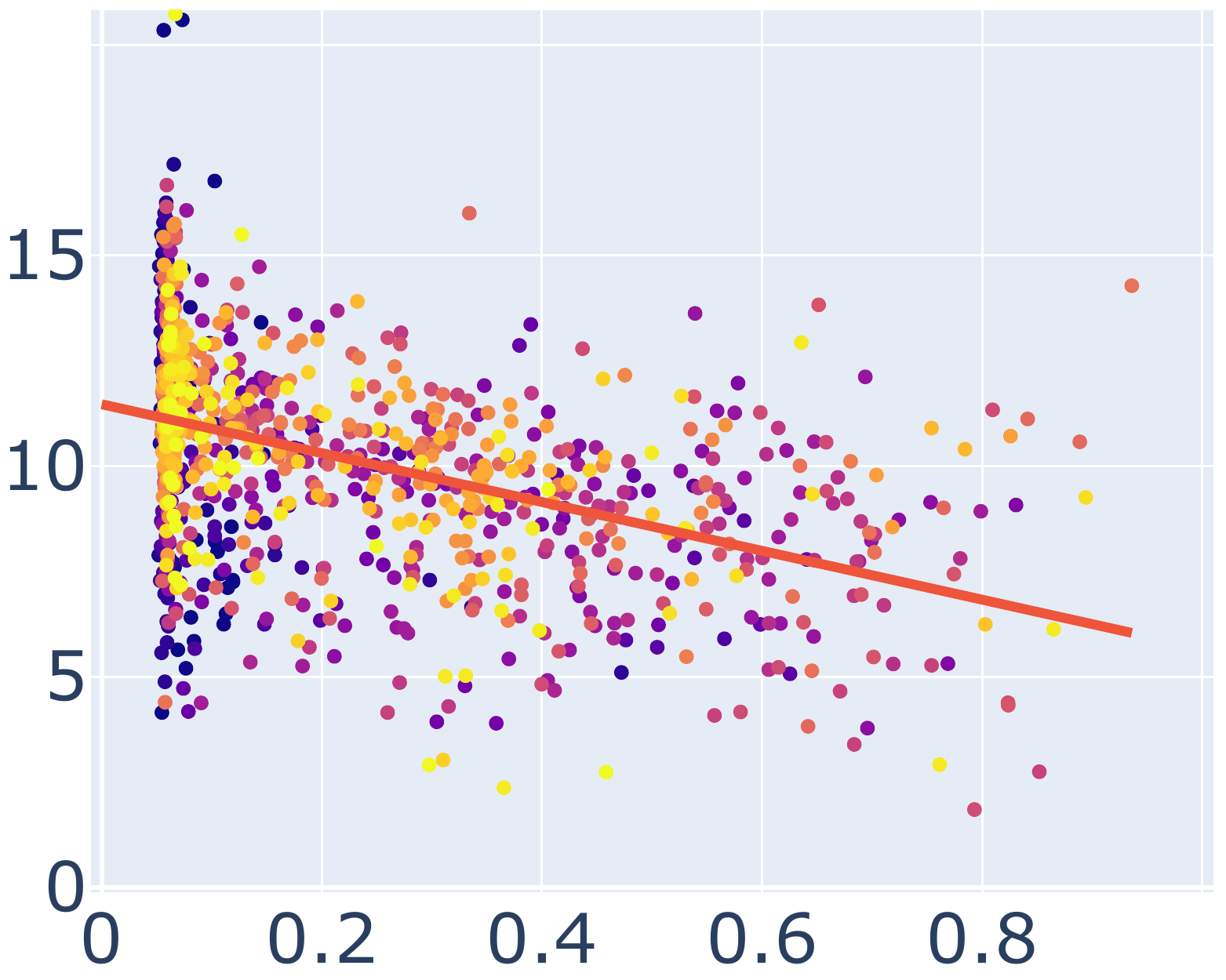}
         \caption{LLaMA first, $\rho=-0.47$}
         \label{fig:y equals x}
     \end{subfigure}    
     \begin{subfigure}[b]{0.23\textwidth}
         \centering
         \includegraphics[width=\textwidth]{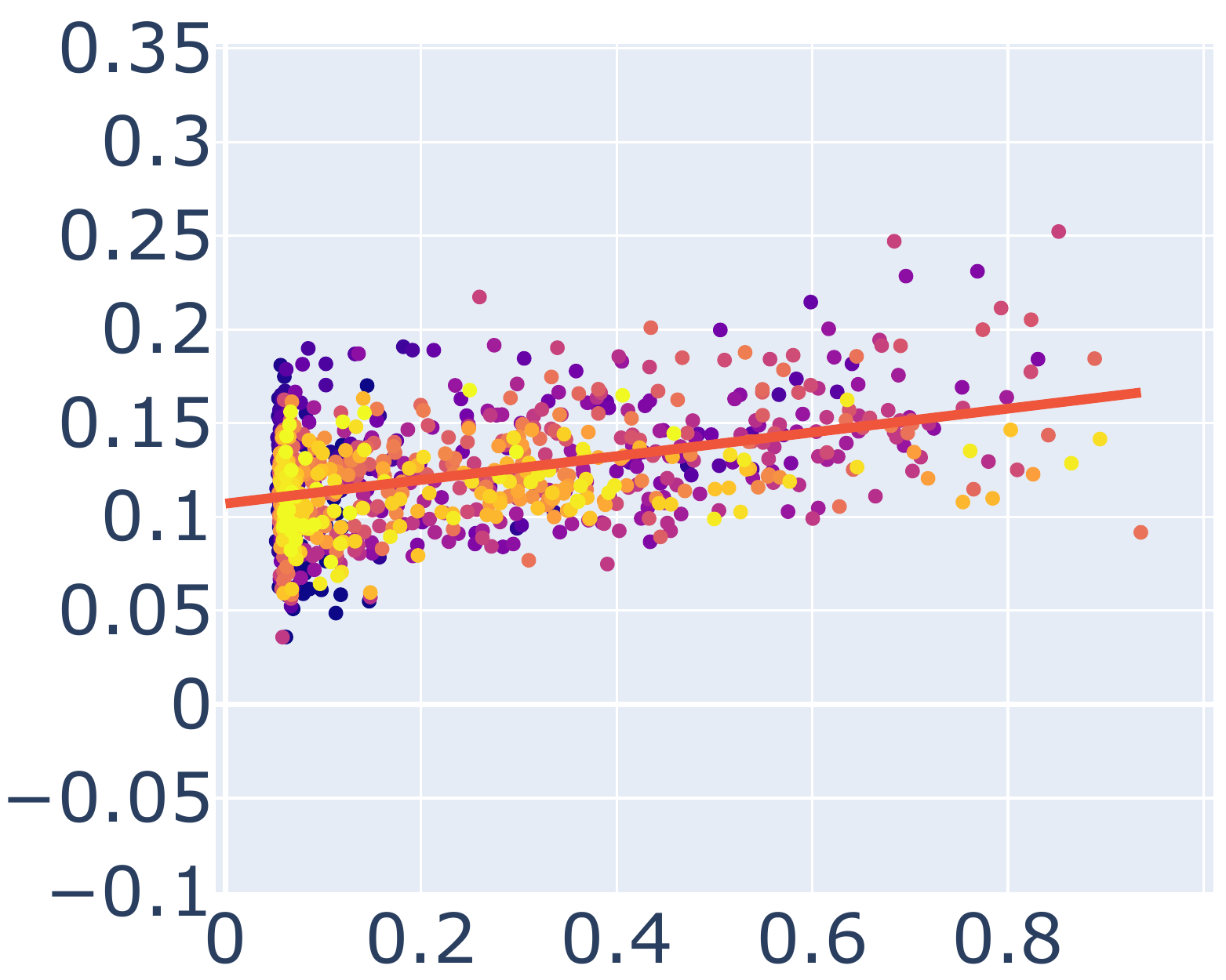}
         \caption{LLaMA last, $\rho=0.44$}
         \label{fig:y equals x}
     \end{subfigure}    
     \begin{subfigure}[b]{0.23\textwidth}
         \centering
         \includegraphics[width=\textwidth]{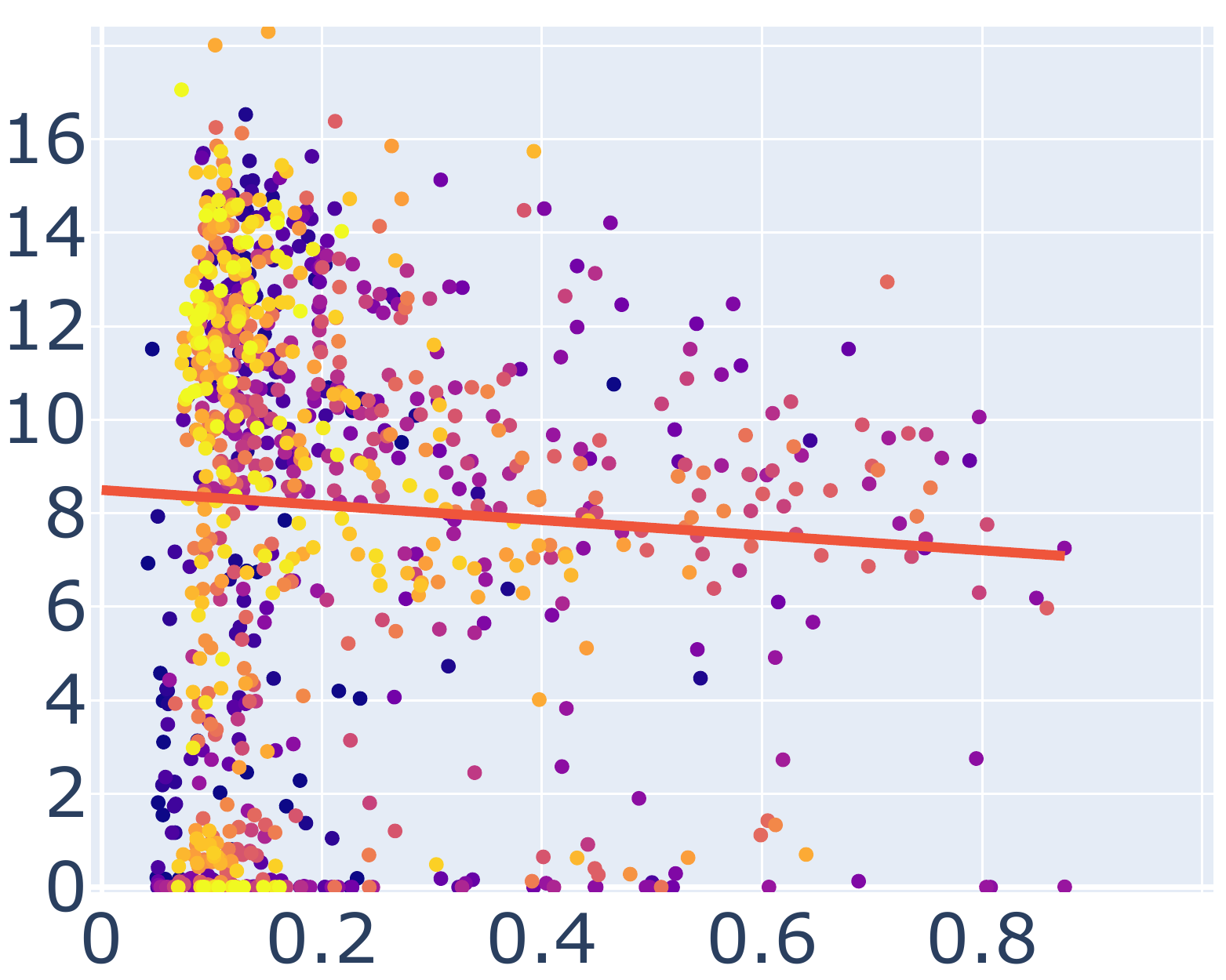}
         \caption{OLMo first, $\rho=-0.03$}
         \label{fig:y equals x}
     \end{subfigure}    
     \begin{subfigure}[b]{0.23\textwidth}
         \centering
         \includegraphics[width=\textwidth]{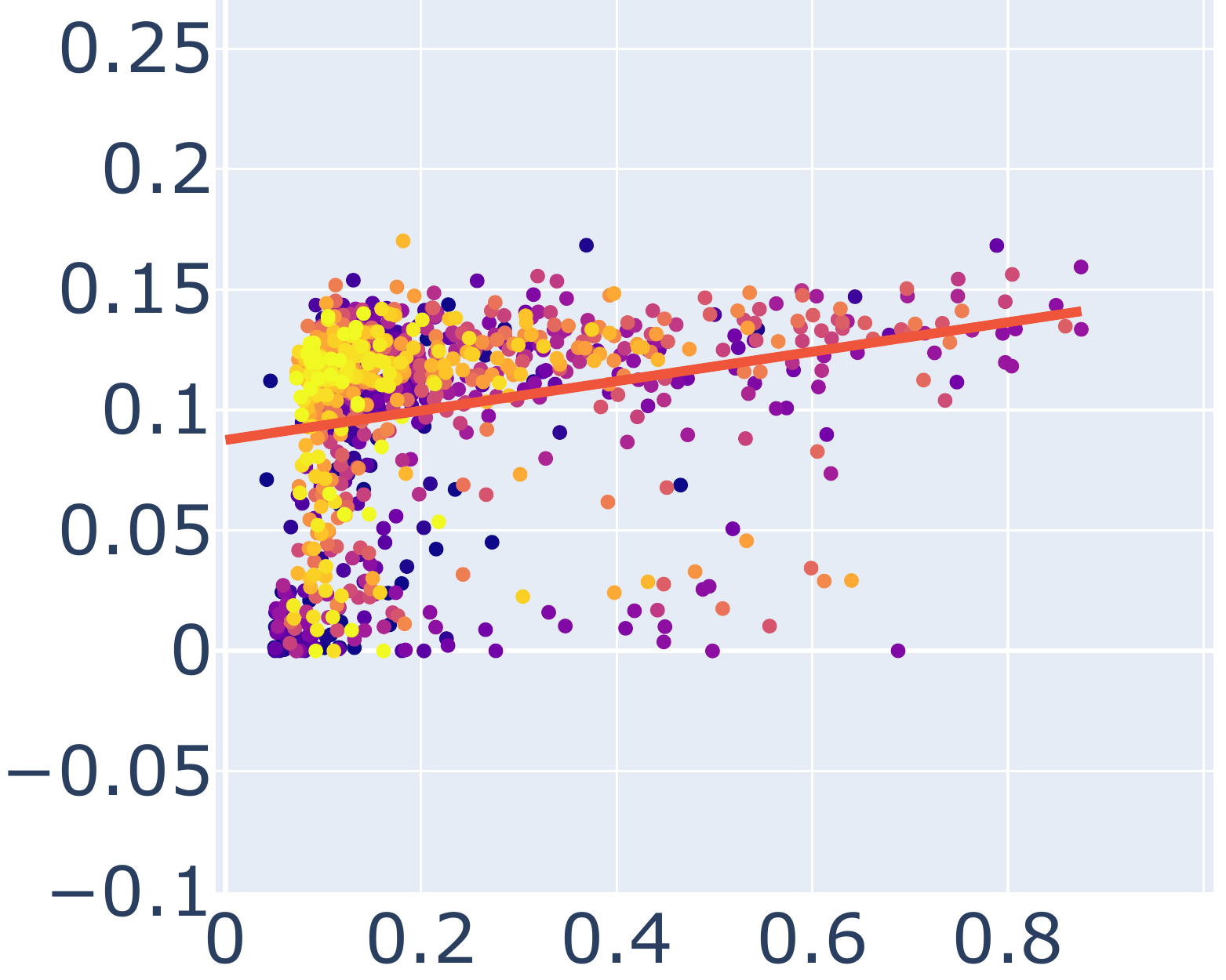}
         \caption{OLMo last, $\rho=0.26$}
         \label{fig:y equals x}
     \end{subfigure}    
     \begin{subfigure}[b]{0.23\textwidth}
         \centering
         \includegraphics[width=\textwidth]{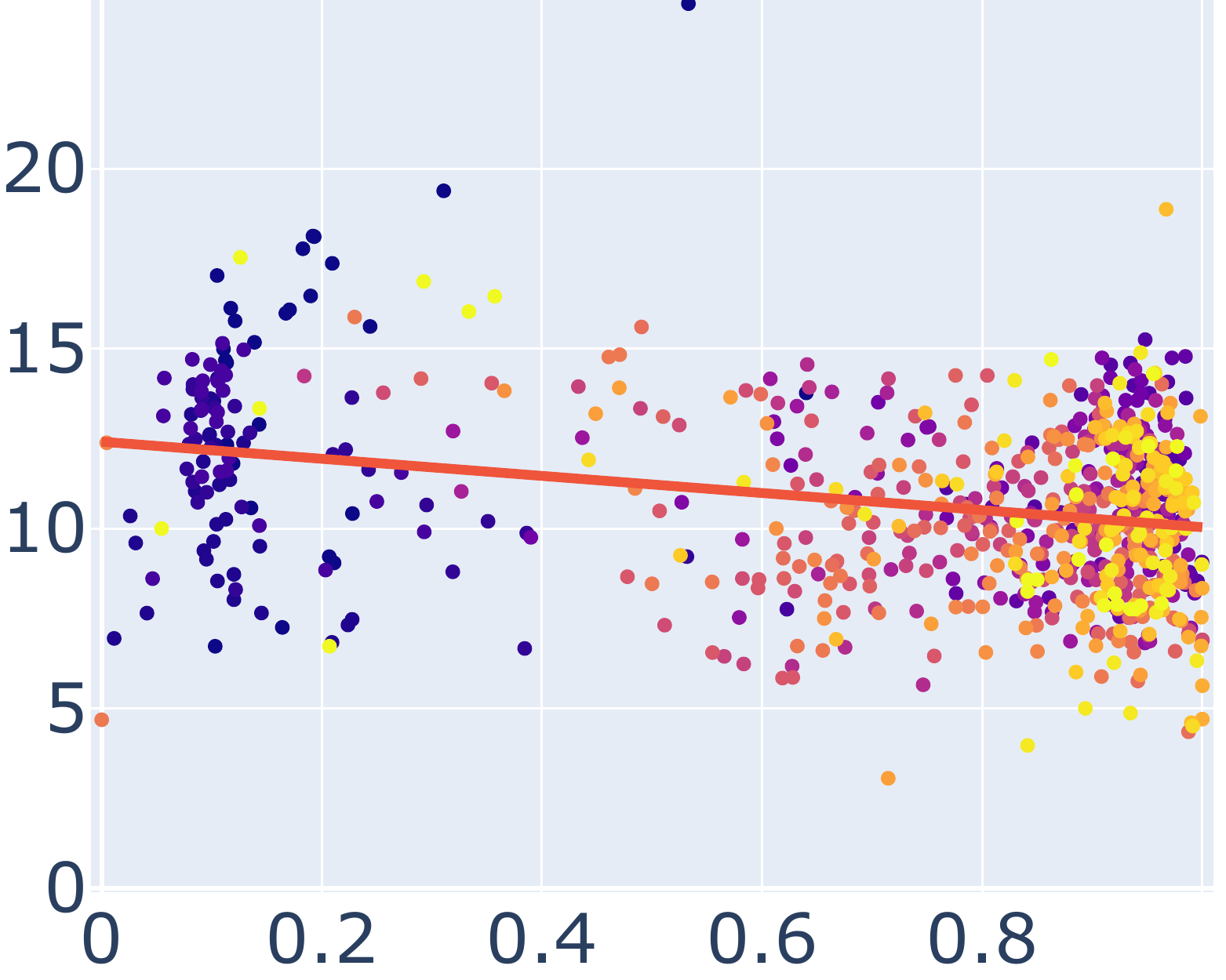}
         \caption{Qwen first, $\rho=-0.27$}
         \label{fig:y equals x}
     \end{subfigure}    
     \begin{subfigure}[b]{0.23\textwidth}
         \centering
         \includegraphics[width=\textwidth]{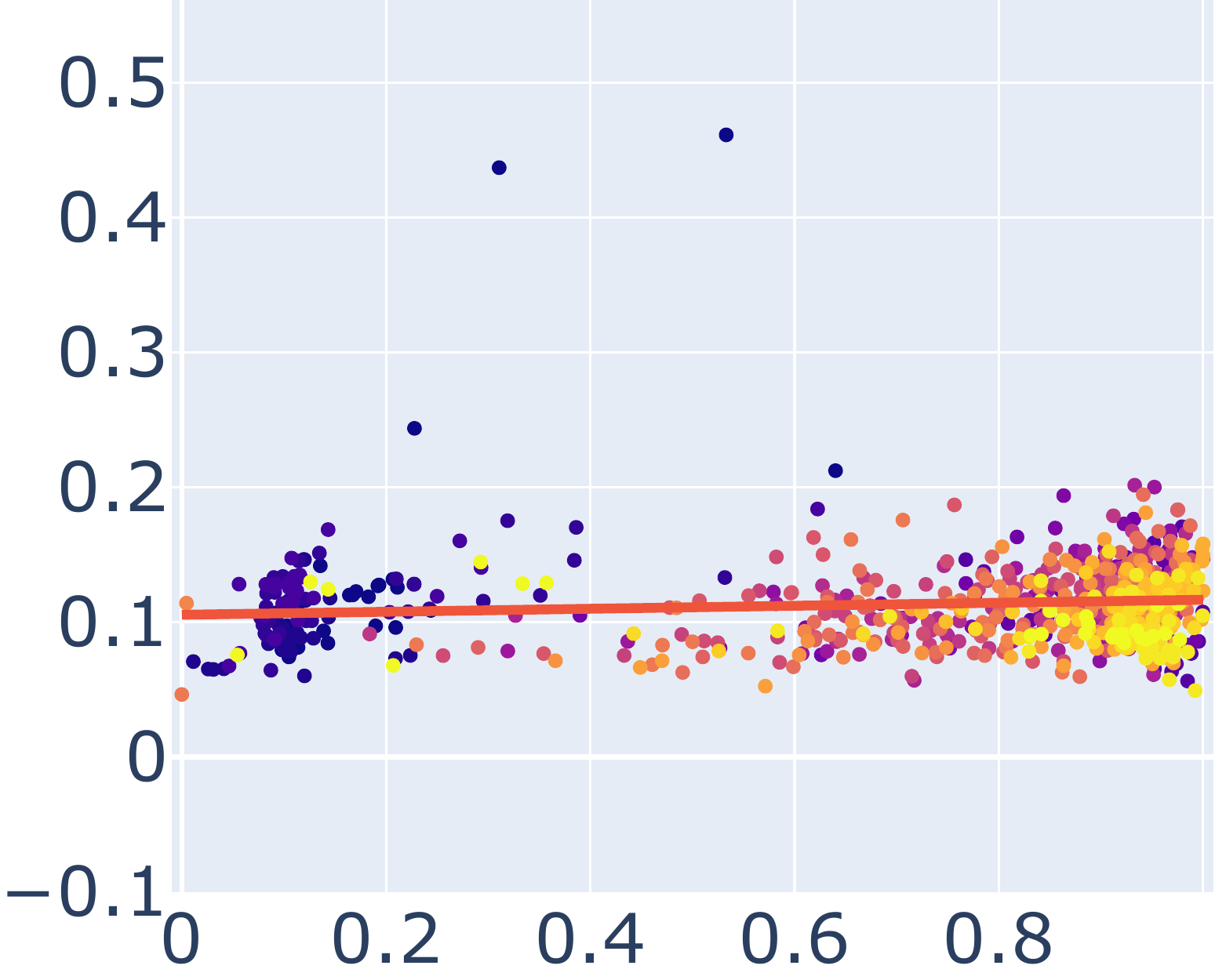}
         \caption{Qwen last, $\rho=0.10$.}
         \label{fig:y equals x}
     \end{subfigure}        
    \caption{The relationship between the L1 norm of rows in query projection matrices (x-axis) and the utility scores of the first or last dimensions (y-axis). Each dot represents an attention head. The lighter dot color represents the deeper layers. The red line represents the linear regressor.}
    \label{fig:attn-vs-weights}
\end{figure}

\begin{figure*}
    \centering
     \begin{subfigure}[b]{0.3\textwidth}
         \centering
         \includegraphics[width=\textwidth]{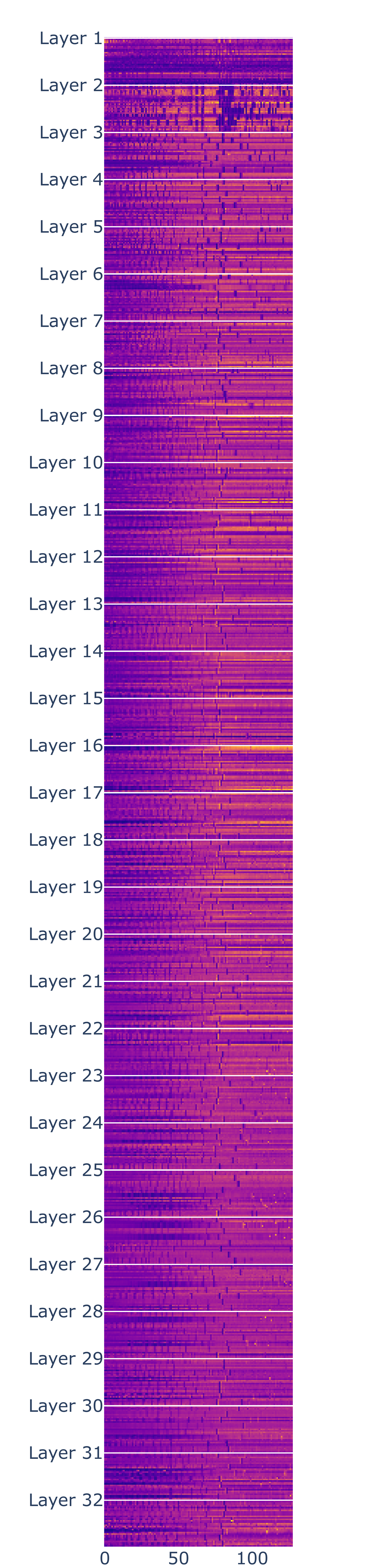}
         \caption{LLaMA}
     \end{subfigure}    
     \begin{subfigure}[b]{0.3\textwidth}
         \centering
         \includegraphics[width=\textwidth]{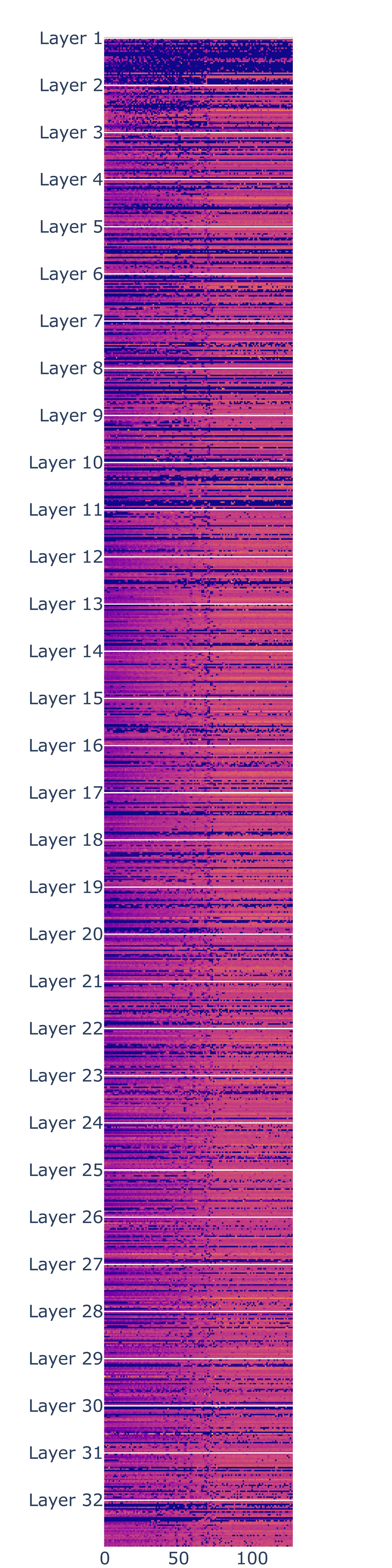}
         \caption{OLMo}
     \end{subfigure}    
     \begin{subfigure}[b]{0.3\textwidth}
         \centering
         \includegraphics[width=\textwidth]{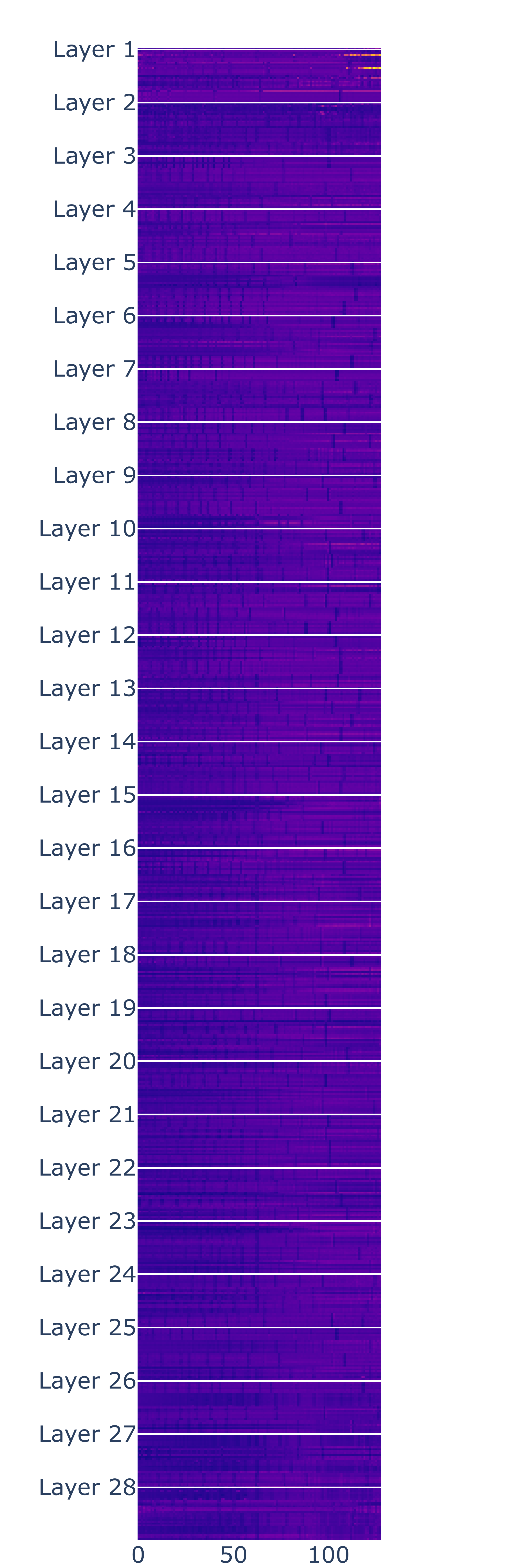}
         \caption{Qwen}
     \end{subfigure}    
    \caption{Visualizing the L1 norm of the rows in the query projection matrices.}
    \label{fig:all-weight}
\end{figure*}

\begin{figure*}
    \centering
     \begin{subfigure}[b]{0.3\textwidth}
         \centering
         \includegraphics[width=\textwidth]{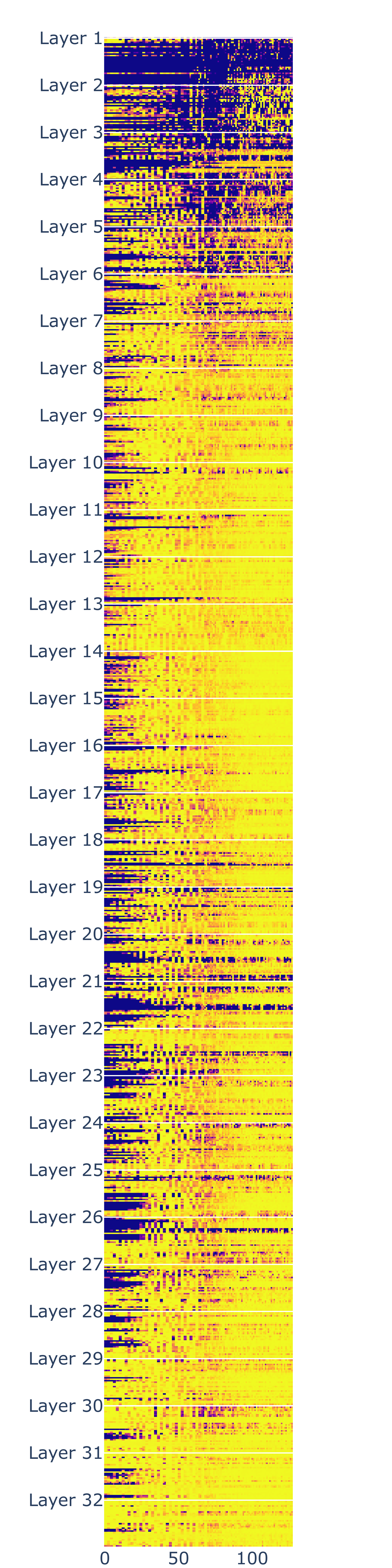}
         \caption{LLaMA}
     \end{subfigure}    
     \begin{subfigure}[b]{0.3\textwidth}
         \centering
         \includegraphics[width=\textwidth]{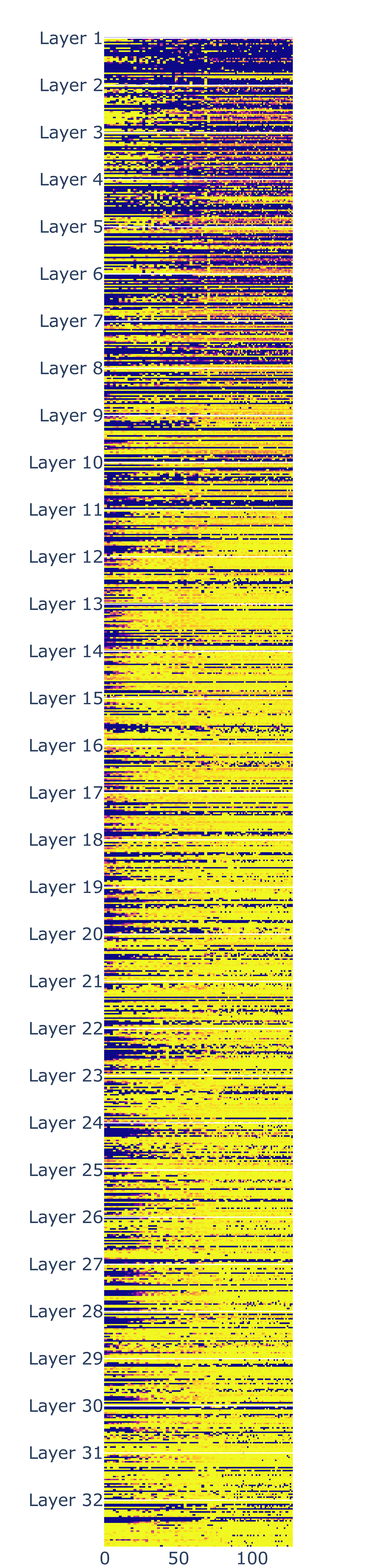}
         \caption{OLMo}
     \end{subfigure}    
     \begin{subfigure}[b]{0.3\textwidth}
         \centering
         \includegraphics[width=\textwidth]{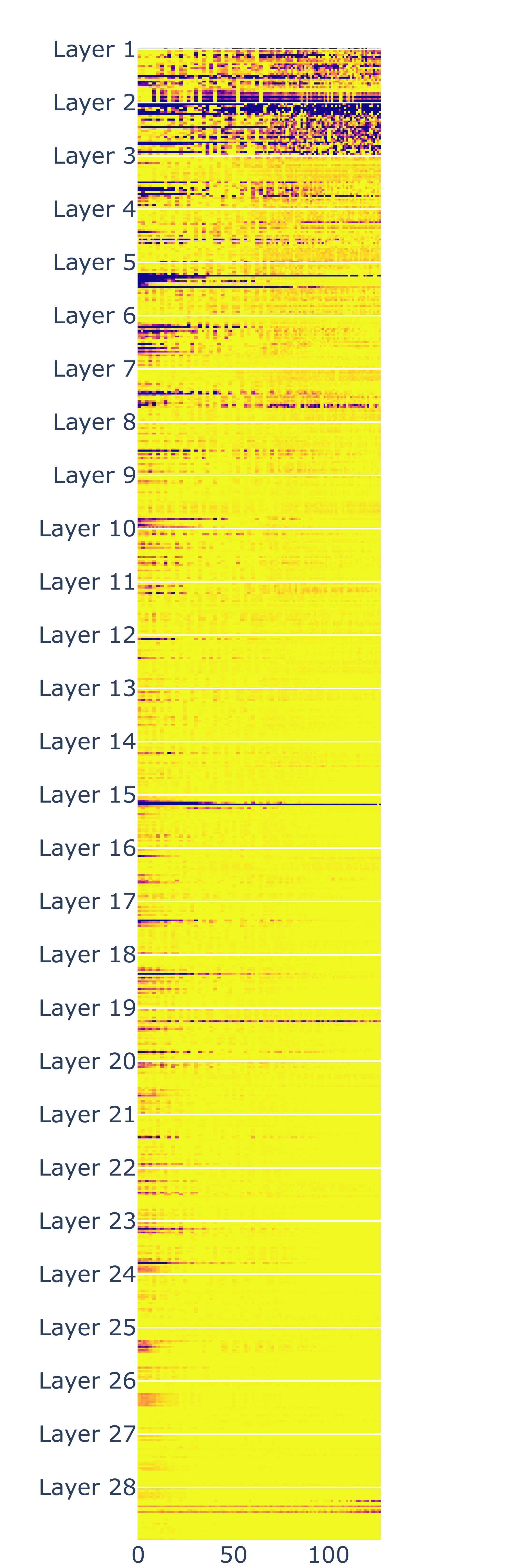}
         \caption{Qwen}
     \end{subfigure}    
    \caption{Visualizing the utilizatio score for the dimensions in the query vectors.}
    \label{fig:all-mask}
\end{figure*}

\end{document}